\documentclass[10pt,twocolumn,letterpaper]{article}

\usepackage{wacv}              

\usepackage{graphicx}
\usepackage{amsmath}
\usepackage{amssymb}
\usepackage{booktabs}
\usepackage{multirow}
\usepackage{comment}
\usepackage{times}
\usepackage{epsfig}
\usepackage[table]{xcolor}

\usepackage{dsfont}
\usepackage{wrapfig}
\usepackage[Export]{adjustbox}
\usepackage{tabularx}
\usepackage{cellspace}
\usepackage[accsupp]{axessibility}
\usepackage{capt-of}

\usepackage[pagebackref,breaklinks,colorlinks]{hyperref}

\usepackage{tikz}
\usepackage[capitalize]{cleveref}
\crefname{section}{Sec.}{Secs.}
\Crefname{section}{Section}{Sections}
\Crefname{table}{Table}{Tables}
\crefname{table}{Tab.}{Tabs.}

\def\wacvPaperID{1610} 
\def\confName{WACV}
\def\confYear{2025}

\newcommand\copyrighttext{%
  \footnotesize \textcopyright 2025 IEEE. Personal use of this material is permitted.
  Permission from IEEE must be obtained for all other uses, in any current or future
  media, including reprinting/republishing this material for advertising or promotional
  purposes, creating new collective works, for resale or redistribution to servers or
  lists, or reuse of any copyrighted component of this work in other works.
  }
\newcommand\copyrightnotice{%
\begin{tikzpicture}[remember picture,overlay]
\node[anchor=south,yshift=10pt] at (current page.south) {\fbox{\parbox{\dimexpr\textwidth-\fboxsep-\fboxrule\relax}{\copyrighttext}}};
\end{tikzpicture}%
}

\begin{document}

\title{Prior2Posterior: Model Prior Correction for Long-Tailed Learning}

\author{S Divakar Bhat* \quad Amit More* \quad  Mudit Soni \quad  Surbhi Agrawal\\
Honda R\&D Co., Ltd. \\Tokyo, Japan\\
}






\twocolumn[{%
\renewcommand\twocolumn[1][]{#1}%
\maketitle
\copyrightnotice
\begin{center}
  \centering
\hspace{1.cm}
\begin{minipage}{0.26\linewidth}
    \includegraphics[trim={4cm 2.cm 4cm 3cm},clip,width=\textwidth]{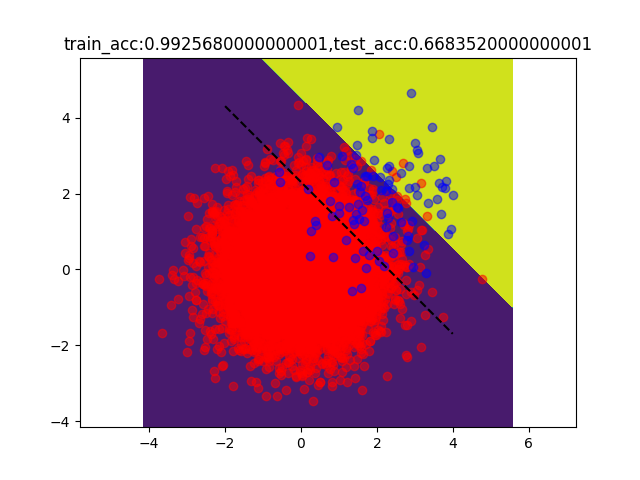}
\end{minipage}
\hfill
\begin{minipage}{0.26\linewidth}
    \includegraphics[trim={4cm 2.cm 4cm 3cm},clip,width=\textwidth]{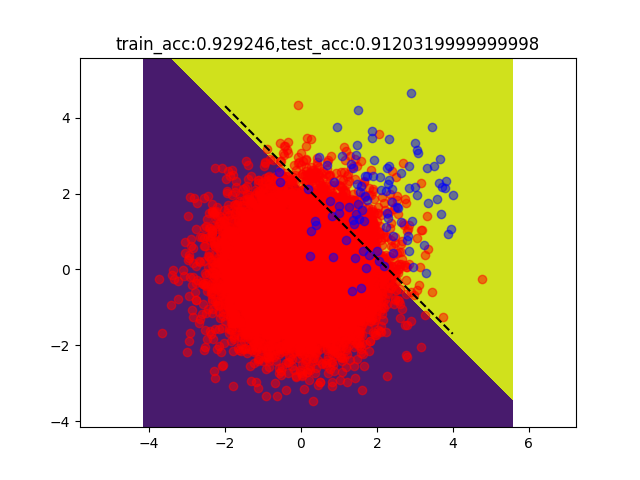}
\end{minipage}
\hfill
\begin{minipage}{0.26\linewidth}
    \includegraphics[trim={4cm 2.cm 4cm 3cm},clip,width=\textwidth]{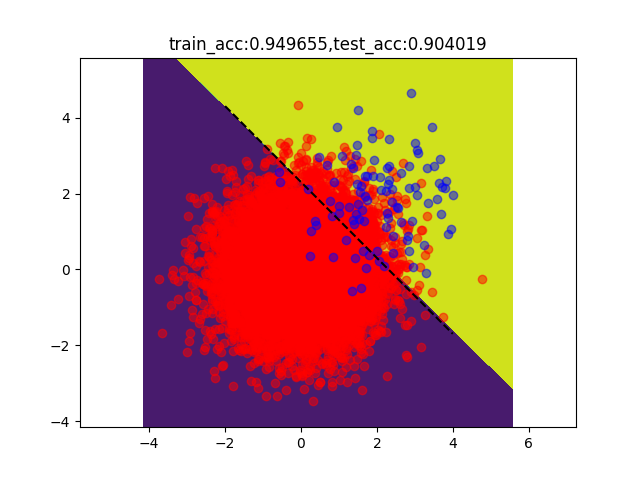}
\end{minipage}        
\hspace{1.cm}
\captionof{figure}{We present the results on a toy dataset with the imbalance factor of $100$. The trained classifier regions are shown in purple and yellow colors and the ideal, Bayes classifier, boundaries are shown as a dashed black line in each figure. 
(a) Classifier boundaries are naturally biased when using a naive cross-entropy (CE) loss.
(b) Using class frequencies for post-hoc correction removes the classifier bias to some extent. 
(c) Using proposed post-hoc correction with learned prior, boundary is adjusted very close to the optimal Bayes' classifier. \label{fig:toy}}
\end{center}
}]

\begin{abstract}
\def\thefootnote{*}\footnotetext{Equal contribution}
Learning-based solutions for long-tailed recognition face difficulties in generalizing on balanced test datasets. 
Due to imbalanced data prior, the learned \textit{a posteriori} distribution is biased toward the most frequent (head) classes, leading to an inferior performance on the least frequent (tail) classes. 
In general, the performance can be improved by removing such a bias by eliminating the effect of imbalanced prior modeled using the number of class samples (frequencies). 
We first observe that the \textit{effective prior} on the classes, learned by the model at the end of the training, can differ from the empirical prior obtained using class frequencies.  
Thus, we propose a novel approach to accurately model the effective prior of a trained model using \textit{a posteriori} probabilities. 
We propose to correct the imbalanced prior by adjusting the predicted \textit{a posteriori} probabilities (Prior2Posterior: P2P) using the calculated prior in a post-hoc manner after the training, and show that it can result in improved model performance. 
We present theoretical analysis showing the optimality of our approach for models trained with naive cross-entropy loss as well as logit adjusted loss. 
Our experiments show that the proposed approach achieves new state-of-the-art (SOTA) on several benchmark datasets from the long-tail literature in the category of logit adjustment methods. 
Further, the proposed approach can be used to inspect any existing method to capture the \textit{effective prior} and remove any residual bias to improve its performance, post-hoc, without model retraining. 
We also show that by using the proposed post-hoc approach, the performance of many existing methods can be improved further. 

\end{abstract}

\begin{figure}[t]
\centering
\includegraphics[width=\columnwidth,trim={0cm, 30cm, 1.0cm, 0cm}, clip]{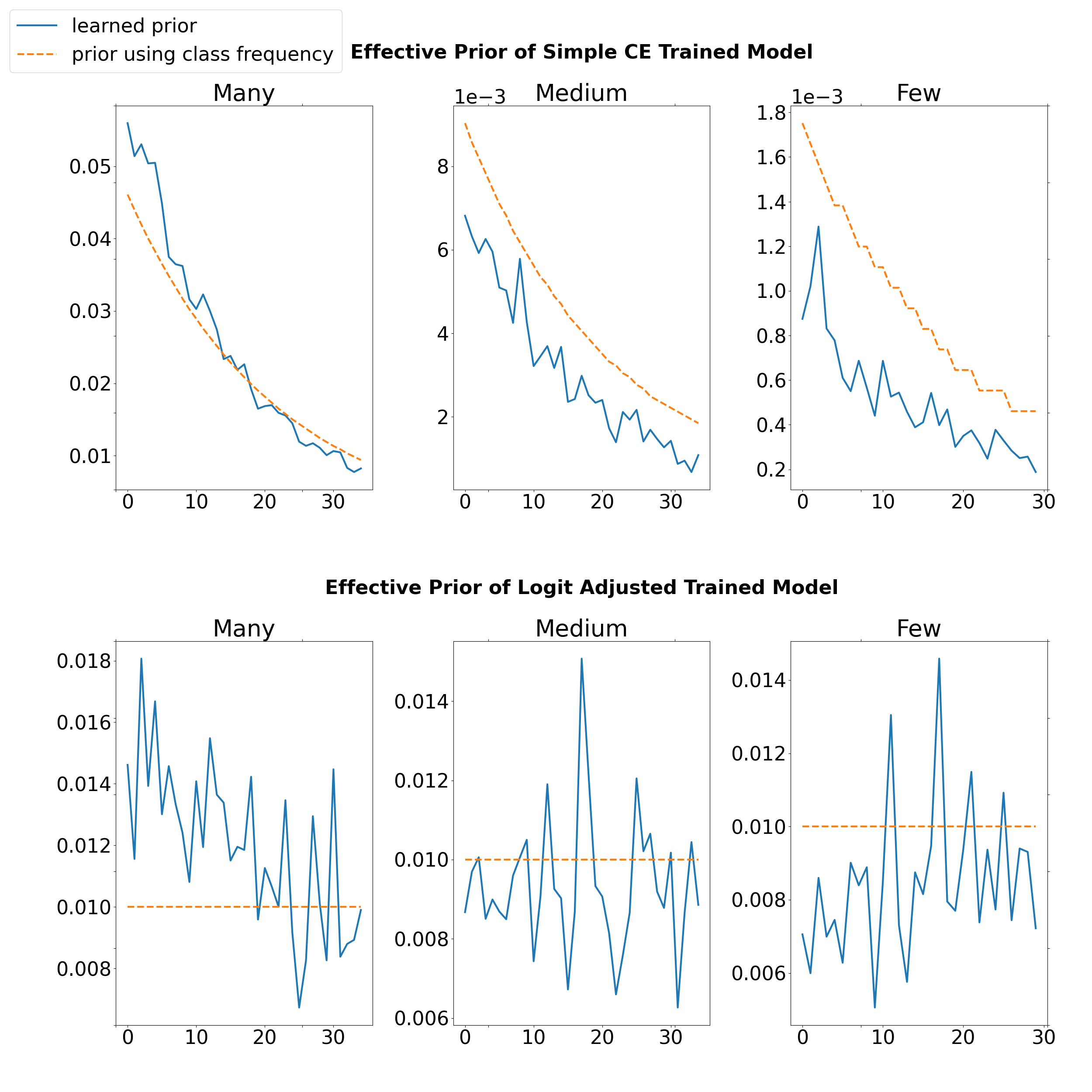}
\caption{ We show marginal class probabilities, $P(y)$, for classes on CIFAR$100$-LT dataset with imbalance factor of $200$. The effective prior calculated using proposed approach and using class frequencies are shown for head classes (Many-first column), tail classes (Few-third column) and rest of the classes (Medium-middle column). 
One may note that model shows bias towards the head classes. 
Further, the bias is under represented by class frequencies for head classes and is over represented for other classes. }
\label{fig:learned_priors}
\end{figure}

\section{Introduction}
One of the primary difficulties in the case of long-tailed recognition is a distribution mismatch between training and test datasets. 
The training dataset is usually dominated by a large number of examples from a few classes called majority or head classes. 
As a result, a naive deep neural network model trained on such a dataset using simple cross-entropy loss is biased toward the dominant classes. 
This bias manifests in the form of higher accuracy on head classes and poor accuracy on tail classes, i.e. accuracy is directly proportional to the number of training examples for a given class. 
One of the early solutions proposed to mitigate the effects of such a data imbalance appears in \cite{margineantu2000does,lawrence2002neural}. 
Generally, the model bias can be removed by artificial re-balancing of the datasets and scaling of loss functions or model prediction scores. 
However, artificial balancing techniques such as under-sampling or oversampling the dataset can affect the quality of the learned features.  

In~\cite{kang2019decoupling}, authors proposed a decoupled training strategy to first learn good quality features and then debias the model by retraining the classifier. 
The authors showed that vanilla empirical loss trained with instance-balanced sampling is enough to generate good-quality features. 
Once the features are learned, the model can be debiased by retraining the classifier with different debiasing techniques such as classifier weight scaling, weight normalization and class-balanced training.
Similarly, \cite{menon2020long,balms} proposed logit adjusted loss and showed that retraining the classifier to introduce a relative margin between pairs of classes leads to improvements. 
Approaches such as these show the effectiveness of the 2-stage training paradigm widely adopted in long-tail learning.  


Despite using different debiasing techniques, we observe that a small bias toward the head classes remains in the final model. 
In this work, we propose to mitigate such a residual bias in a post-hoc manner. 
In \cite{menon2020long}, authors proposed post-hoc logit adjustment when the model is trained with plain cross-entropy loss. 
The bias is modeled using the number of samples per class (also called class frequencies), which can re-adjust the classifier boundaries after training. 
However, we argue that class frequencies can not accurately represent the bias present in the trained model. 
Figure~\ref{fig:learned_priors} shows the effective prior learned by a DNN model when trained on the CIFAR100-LT dataset with an imbalance factor of $200$. 
As shown in the figure, the model bias or a prior toward the head classes is under-represented by class frequencies, i.e. model has a stronger bias than what is indicated by class frequencies. 
Similarly, model bias is smaller than what is implied by class frequencies for other classes. 
One may attribute such behavior to the over-fitting nature of DNN models. 
In general, DNN model output probabilities can be considered as an estimate of the Bayesian \textit{a posteriori} probabilities \cite{richard1991neural}. 
However, the accuracy of these estimates depend on network complexity, the amount of available training data and the effectiveness of the training algorithm i.e. convergence to the global minimum~\cite{richard1991neural,lawrence2002neural}. 
Nonetheless, using class frequencies for post-hoc debiasing will generally lead to sub-optimal improvements. 
Further, this approach is valid only when the model is trained with plain cross-entropy loss and instance-balanced data sampling, and combining it with complex training approaches is non-trivial. 
Thus, we propose to model the residual bias (also referred to as a learned prior) accurately using \textit{a posteriori} class probabilities of a trained model and remove the bias in a post-hoc manner. 
Despite being simple, we show that the proposed approach is very effective, leads to superior performance, and can rival many state-of-the-art methods. 

The proposed post-hoc adjustment is statistically consistent with many training paradigms, unlike class frequency-based post-hoc adjustment, and hence can be applied to existing logit adjustment losses as well. 
We apply the proposed post-hoc adjustment to many existing methods without retraining the model and show that it can boost the model performance further. 


We summarize the specific contributions of our work:
\begin{enumerate}
\item We show that the effective prior of a model differs from class frequency based prior and propose how to calculate such a prior. 
\item We present post-hoc adjustment to remove the model bias using calculated prior and prove the theoretical optimality.  
\item We validate our claims on several datasets, including CIFAR10-LT \& CIFAR100-LT with different imbalances, ImageNet-LT and iNaturalist18 achieving new SOTA in the category of logit adjustment based approaches. 
\item The proposed approach can be applied to existing methods, removing the residual bias and improving their performance further without model retraining. 
\end{enumerate}

\section{Related Work}
\textbf{Data re-balancing:} 
Early studies on the effect of data imbalance on a learned model can be found in~\cite{margineantu2000does,lawrence2002neural}. 
Generally, the dataset is artificially balanced by either oversampling rare classes~\cite{buda2018systematic} or under-sampling head classes~\cite{mani2003knn}. 
To mitigate limitations of these methods generative approaches are used to generate samples in feature~\cite{li2022long} and image space~\cite{du2023global}. 
\\
\textbf{Loss modification:} 
Engineering the loss functions can be one natural way to tackle the class imbalance problem. 
In line with this~\cite{lin2017focal} proposed a novel loss function to assign a higher loss value to poorly classified samples. 
As the tail class examples often show high misclassification rates, re-scaling the loss using inverse class frequencies was suggested by~\cite{huang2016learning,wang2017learning}, whereas~\cite{cui2019class} proposed to eliminate class imbalance by using the concept of effective number of samples. 
The gradient modification is proposed in~\cite{wang2021adaptive,wang2021seesaw} to address class imbalance. 
In~\cite{bhat2023robust} authors proposed to boost the gradients by artificially suppressing the prediction scores for effective learning. 
\\
\textbf{Distribution alignment:} 
Long-tailed recognition problem is posed as a distribution misalignment problem in~\cite{cao2019learning} and it is proposed to use class dependent margin to modify the decision boundaries. 
Following this, \cite{balms,hong2021disentangling,menon2020long} tried to reduce the gap between the learned posterior and the uniform distribution.  
In~\cite{wang2023margin,zhang2021distribution} authors propose to learn additional logit calibration layer to mitigate bias. 
Our approach is similar to these methods in its spirit, however we propose to scale probabilities (adjust the logit score) based on effective prior (imbalance) learned by the model and hence it is more effective. 
\\
\textbf{Other approaches:} 
\cite{kang2019decoupling} proposed to independently learn the feature extractor and classifier. 
They concretely demonstrate the effectiveness of stage-wise training in long-tailed recognition. 
Performance improvement with careful weight decay tuning is shown in~\cite{alshammari2022long}. 
Long-tailed learning is analyzed from an optimization perspective in \cite{rangwani2022escaping,ma2023curvature,zhou2023class} and improves the generalization on tail classes. 
\section{Preliminaries}

\subsection{Problem Formulation}
Consider a $\mathcal{C}$ class classification problem where the objective is to learn a mapping from data instances $\mathcal{X} = \{x_1, x_2, ..., x_N\}$ to the corresponding ground-truth labels, $\mathcal{Y} = \{ y_1, y_2, ..., y_N\}$ where $N$ denotes total number of samples. 
Further, let $n_i$ denote the number of samples for $i^{th}$ class, such that $n_i \neq n_{j}$. 
Let $P(y|x)$ represent probability of a predicted class conditioned on a given data sample $x$. 
Further, let $P(y)$ represent the prior probability distribution for labels over given dataset. 
Usually, we train a model to predict an unnormalized class probability scores, logits denoted by $z$, using a deep neural network represented by $f(x)$ and the probabilities $P(y|x)$ are calculated by normalizing the logits using a softmax function. 

In case of a long-tailed recognition, the training set is imbalanced, however, the goal is to maximize the recognition performance over a balanced test set. 
We denote predicted conditional and marginal class probabilities over given balanced test set by $P^{t}(y|x)$ and $P^{t}(y)$, respectively. 
Thus the objective is to model $P^{t}(y|x)$ using imbalanced training data, i.e. $P(y|x)$. 
Since training data distribution is highly skewed, the marginal class distributions are not equal for train and test datasets, i.e. $P^{t}(y) \neq P(y)$. 
As a consequence, the learned distribution $P(y|x)$ will be different from what is required, i.e. $P^{t}(y|x)$. 
\subsection{Logit Adjustment}
From Bayes' theorem we have,
\begin{align}
P(y|x) &= \frac{P(x|y)P(y)}{P(x)} \label{eqn:train_bayes} \\
P^{t}(y|x) &= \frac{P^{t}(x|y)P^{t}(y)}{P^{t}(x)} \label{eqn:test_bayes}
\end{align}
In general, the training and the test dataset distributions differ only in the number of samples available per class and it is safe to assume that the generative distributions are the same, i.e. $P(x|y) = P^{t}(x|y)$. 
Thus, simplifying the Eq.~\ref{eqn:train_bayes} and Eq.~\ref{eqn:test_bayes} gives 

\begin{align}
P^t(y|x) = P(y|x) \frac{P^t(y)}{P(y)} \frac{P(x)}{P^t(x)} \label{eqn:ideal_adj}
\end{align}

The conditional distribution on the train data can be adjusted as shown above to get the desired distribution on the test dataset. 
In practice, the scaling by $\frac{P(x)}{P^t(x)}$ is inconsequential as it can be absorbed into the normalization process. 
The term $\frac{P^t(y)}{P(y)}$ represents the actual correction factor to mitigate the model bias. 
Above equation suggests that appropriate posterior distribution on the balanced test set can be estimated by simply adjusting the predicted logit scores as,
\begin{align}\label{eqn:prior-adj}
z^{t} = z - \log P(y) + \log P^{t}(y). 
\end{align}
This simple but very powerful idea underlies behind all logit adjustment approaches. 
However, it requires that training time data distribution be known, \textit{a priori}. 
Usually marginal distribution $P(y)$ is approximated using class frequencies~\cite{menon2020long,balms}. 
\cite{menon2020long} proposed to use instead a tuned version of class frequencies leading to  
\begin{align}\label{eqn:prior-comp_}
z_i^{t} = z_i - \alpha \log \frac{n_i}{\sum_k n_k}  + \log P^{t}(y_i)
\end{align}
where, $\alpha$ is tuned with the help of holdout validation set. 

From similar inspirations, \cite{menon2020long,balms} have also employed a modified loss function as part of a decoupled training process in an effort to directly model $P^{t}(y|x)$. This modified softmax cross entropy loss can be written as, 
\begin{align}\label{eqn:logit_adj_loss}
\mathcal{L}(f(x), y) =& -\log \frac{e^{z_i + \alpha \log P(y_i)}}{\sum_{k}^{\mathcal{C}} e^{z_k + \alpha \log P(y_k)}}
\end{align}
where, $P(y_i)$ is estimated using class frequencies as in Eq.~\ref{eqn:prior-comp_} and $z_i$ denotes the logit score corresponding to the label $i$ for the sample $x$.
Approximating class prior $P(y)$ using sample frequencies, although simple and effective, comes with a strong assumption that conditional distribution $P(y|x)$ is learned by the model accurately. 
Given the over-fitting nature of deep neural networks and imperfections of stochastic optimization algorithms, the learned distribution does not always represent the true \textit{a posteriori} distribution~\cite{richard1991neural,lawrence2002neural,guo2017calibration,szegedy2016rethinking}. 
As a result using class frequencies for logit adjustment can be sub-optimal. 

In the subsequent sections, we introduce our post-hoc logit adjustment approaches using proposed novel model prior correction method. 
First we present the analysis on the models trained with plain CE loss. 
Next we generalize this approach to train time logit adjustment approaches in a two stage decoupled training framework. 
We propose to calculate the effective prior or a bias of the model using its predictions. 
We argue that calculating class prior using conditional probabilities predicted by the model for logit adjustment is more appropriate. 
This is natural, since the effective prior and the learned posterior distributions are interdependent. 
If logits used for class prediction represent approximate distribution learned by a limited capacity model, then it is appropriate to derive underlying \textit{effective} prior using model prediction itself rather than class frequencies. 



\section{Proposed Approach}
Ideally when $P(y|x)$ is modeled correctly, the adjustment shown in Eq.~\ref{eqn:ideal_adj} is optimal in the sense that it has the desired effective prior $P^t(y)$. 
However, in practice, $P(y|x)$ is modeled by a DNN which generally tend to overfit on the head classes and underfit on the tail classes when training dataset is imbalanced. 
We represent the learned a posteriori distribution by a model as $P^m(y|x)$ which approximates the training distribution $P(y|x)$. 
We can generalize the idea of logit adjustment and denote the adjusted version of $P^m(y|x)$ by $P^a(y|x)$ and formally define what constitutes an optimal adjustment based on Eq.~\ref{eqn:ideal_adj}. 
\subsubsection*{\textbf{Definition:}}The adjusted distribution $P^a(y|x)$ is optimal for the test dataset with marginal distributions $P^t(x)$ and $P^t(y)$ if it satisfies following property
\begin{align}
P^a(y) = \int P^a(y|x) P^t(x) dx = P^t(y) \label{eq:opt_def}
\end{align}

It is easy to see that in an ideal case when $P^m(y|x) = P(y|x)$, the adjusted distribution given by Eq.~\ref{eqn:ideal_adj} is an optimal adjustment. 

\subsection{Post-hoc correction with learned prior}
We now show the optimal adjustment for $P^m(y|x)$ when model is trained with plain cross-entropy loss. 
\subsubsection*{\textbf{Theorem 1.}}Let $P^m(y|x)$ be the posterior distribution learned by the model, then the optimal adjustment to match the test distribution prior is given by,
\begin{align}
P^a(y|x) &= P^m(y|x)\frac{P^t(y)}{P^m(y)}\frac{P(x)}{P^t(x)} \label{eq:pa}
\end{align}
where, 
\begin{align}
P^m(y)= \int P^m(y|x) P(x) dx \label{eq:pm}
\end{align}

$P^m(y)$ is the effective prior on the training dataset learned by the model. 
\\\textbf{Proof.} 
The effective prior of the adjusted probabilities on the test distribution is given by,
\begin{align}
P^a(y) &= \int P^a(y|x) P^t(x) dx\\
    &= \int P^m(y|x)\frac{P^t(y)}{P^m(y)}\frac{P(x)}{P^t(x)} P^t(x) dx\\
    &= \frac{P^t(y)}{P^m(y)} \int P^m(y|x)P(x) dx\\
    &= P^t(y)
\end{align}

Thus the adjusted distribution $P^a(y|x)$ represents the same prior as required by the test dataset. It should be noted that, under ideal conditions when $P^m(y|x) = P(y|x)$, Eq. \ref{eq:pa} reduces to the class frequency based adjustment. 

\begin{table*}[]
\centering
\addtolength{\tabcolsep}{0.7em}
\begin{tabular}{@{}l|ccc|ccc@{}}
\toprule
Dataset          & \multicolumn{3}{c}{CIFAR10-LT} & \multicolumn{3}{c}{CIFAR100-LT} \\ \midrule
Imbalance factor & 200       & 100      & 10      & 200       & 100       & 10      \\ \midrule
Cross Entropy (CE) baseline &   77.02&          82.31&         91.87&           43.39&           47.58&        64.34   \\

\midrule 
Balms~\cite{balms} \footnotesize NeurIPS'20 & 81.50& 84.90& 91.30& 45.50& 50.80& 63.00 \\ 
$\text{LDAM-DRW}^{\dagger}$~\cite{cao2019learning} \footnotesize NeurIPS'19& -& 77.03& 88.16& -& 42.04& 58.71 \\
MARC~\cite{wang2023margin} \footnotesize ACML'23    & 81.10       & 85.30      & -      & 47.40       & 50.80       & -      \\

\midrule 
$\text{CB-CE}^{*}$~\cite{cui2019class} \footnotesize CVPR'19& 68.89& 74.57& 87.49& 36.23& 39.60& 57.90 \\
$\text{cRT}^{+}$~\cite{kang2019decoupling} \footnotesize ICLR'20 & 76.60 & 82.00 & 91.00 & 44.50 & 50.00 & 63.30 \\

\midrule 
AREA~\cite{chen2023area} \footnotesize ICCV'23 & 74.99& 78.88& 88.71& 43.85& 48.83& 60.77	\\
CC-SAM~\cite{zhou2023class} \footnotesize CVPR'23 & 80.94 &83.92 & - & 45.66 & 50.83& -\\
GCL+CR~\cite{ma2023curvature} \footnotesize CVPR'23 &79.90 &83.50 & - & 45.60 &49.80& -\\





\midrule \multicolumn{7}{c}{\cellcolor[gray]{.9} Post-hoc on Stage 1} \\ \midrule
CE + post-hoc adj. with class freq. &           		78.29&          83.82&         92.61&           48.68&           52.93&       65.85 \\
CE + P2P   &           79.43&           84.14&         92.82&         49.61&         53.42&    65.95   \\ 
\midrule \multicolumn{7}{c}{\cellcolor[gray]{.9}Post-hoc on Stage 2} \\  \midrule
CL &           77.16&          84.43&         92.71&           51.67&           56.37&        66.00 \\
CL + P2P    &           81.14&          85.79&         92.94&           51.90&           56.49&      66.00   \\
FT &           80.75&          86.01&         93.18&           52.71&           57.37&         68.12 \\
FT + P2P     &           \textbf{85.77}&          \textbf{87.26}&         \textbf{93.45}&           \textbf{52.92}&           \textbf{57.39}&       \textbf{68.15}  \\ \bottomrule
\end{tabular}
\caption{Top 1 accuracy for CIFAR10-LT and CIFAR100-LT across different imbalance factors. * indicates results reported in~\cite{shu2019meta}.  $\dagger$ indicates results reported in~\cite{cao2019learning}. + indicates results reported in~\cite{balms}. 
}
\label{tab:cifar}
\end{table*}
\begin{table}[]
    \centering
    \addtolength{\tabcolsep}{-0.4em}
\renewcommand{\arraystretch}{1.}
    \begin{tabular}{@{}l|c|c}
\toprule
 Dataset &ImageNet-LT &iNaturalist-18\\ \midrule
Cross Entropy (CE) baseline                                                     & 48.63 & 66.03\\

\midrule 
Balms~\cite{balms} \footnotesize NeurIPS'20     & 52.30 & 70.60 \\
LADE~\cite{hong2021disentangling} \footnotesize CVPR'21        & 53.00 & 70.00 \\
DisAlign~\cite{zhang2021distribution} \footnotesize CVPR'21        & 53.40 & 70.60 \\
MARC~\cite{wang2023margin} \footnotesize ACML'23        & 52.30 & 70.40 \\

\midrule 
cRT~\cite{kang2019decoupling} \footnotesize ICLR'20                        & 49.50 & 67.60  \\
DRO-LT ~\cite{samuel2021distributional} \footnotesize ICCV'21        & 53.50 & 69.70 \\
WB+MaxNorm~\cite{alshammari2022long} \footnotesize CVPR'22        & 53.90 & 70.20 \\

\midrule 
ResLT~\cite{cui2022reslt} \footnotesize TPAMI'22       & 52.90 & 70.20 \\
RBL~\cite{peifeng2023feature} \footnotesize ICML'23        & 53.30 & 70.10 \\
SWA+SRepr~\cite{nam2023decoupled} \footnotesize ICLR'23	& - &70.79   \\


\midrule \multicolumn{3}{c}{\cellcolor[gray]{.9}Ours} \\ \midrule
CE + P2P                                                    & 53.24 & 71.15 \\
CL                                                     & 53.23 & 70.81  \\
CL + P2P                                                  & 53.57 & 71.43 \\
FT                                                         & 53.82 & 71.12 \\
FT + P2P                                                    & \textbf{54.67} & \textbf{71.78}  \\\bottomrule

\end{tabular}
\caption{Top 1 accuracy on ImageNet-LT and iNaturalist-18.}
\label{tab:img-inat-results}
\vspace{-2em}
\end{table}

\begin{table*}[ht]
\centering
\subfloat[]{
\begin{tabular}{@{}l|c|c|c@{}}
\toprule
\multicolumn{4}{c}{ImageNet-LT} \\ \midrule
Method                                   &Arch.       &Orig.      &P2P Adjustment     \\  \midrule
cRT~\cite{kang2019decoupling}  \footnotesize ICLR'20               &x50      	&49.63     &\textbf{50.09}  ($\uparrow$ 0.46) \\
MisLAS~\cite{zhong2021improving} \footnotesize CVPR'21                &r50     &52.74    &\textbf{53.09}  ($\uparrow$ 0.35) \\
RIDE~\cite{wang2020long} \footnotesize ICLR'21                &x50      	&55.69      &\textbf{55.96}  ($\uparrow$ 0.27) \\
BCL~\cite{zhu2022balanced} \footnotesize CVPR'22                &x50       &57.78    &\textbf{58.10}  ($\uparrow$ 0.32) \\
GLMC+BS~\cite{du2023global} \footnotesize CVPR'23                &x50      	&57.19    &\textbf{57.41}  ($\uparrow$ 0.22) \\
GLMC~\cite{du2023global} \footnotesize CVPR'23                &x50      	&56.29    &\textbf{57.77}  ($\uparrow$ 1.48) \\ \bottomrule
\end{tabular}
}
\hspace{0.1em}
\subfloat[]{

\centering
\begin{tabular}{@{}l|c|c@{}}
\toprule
\multicolumn{3}{c}{iNaturalist18} \\ \midrule
Method                                    &Orig.       &P2P Adjustment     \\  \midrule
cRT~\cite{kang2019decoupling} \footnotesize ICLR'20                   &71.22   &\textbf{72.55} ($\uparrow$ 1.33) \\
MisLAS~\cite{zhong2021improving} \footnotesize CVPR'21               &71.48    &\textbf{71.96} ($\uparrow$ 0.48) \\
LWS~\cite{kang2019decoupling} \footnotesize ICLR'20                  &72.04   &\textbf{72.98} ($\uparrow$ 0.94) \\
MetaSAug~\cite{li2021metasaug}  \footnotesize CVPR'21               &69.44   &\textbf{70.29} ($\uparrow$ 0.85) \\

CE+DRW+CMO~\cite{park2022majority} \footnotesize CVPR'22                &70.89 &\textbf{71.30} ($\uparrow$ 0.41) \\
ResLT~\cite{cui2022reslt} \footnotesize TPAMI'22               	&68.48    &\textbf{70.15} ($\uparrow$ 1.67) \\ \bottomrule
\end{tabular}
}
    
\caption{ The performance boost on ImageNet-LT (a) and iNaturalist18 (b) datasets for different methods with proposed approach.}
\label{tab:img_inat-sota}
\end{table*}

\subsection{Logit adjusted Training}

One possible way to mitigate the bias of the model is to correct the probabilities during training time itself and get an unbiased estimator. 
Rearranging the Eq.~\ref{eqn:ideal_adj} gives us, 

\begin{align}
P(y|x) = P^{t}(y|x) \frac{P(y)}{P^{t}(y)} \frac{P^t(x)}{P(x)} \label{eq:ideal_adj2}
\end{align}
We employ a two stage decoupled training for this approach with stage 1 being a simple softmax cross entropy training. 
For stage 2 we use loss function shown in Eq.~\ref{eqn:logit_adj_loss}. 
In this framework, $P^t(y|x)$ is modeled by a DNN and the prediction probabilities are adjusted during training to match $P(y|x)$. 
During inference time, the DNN output is directly used to model $P^t(y|x)$ without any adjustment. 
As model probabilities are already adjusted during training, when the logit adjustment is removed during inference time, ideally model should not show any bias. 
However, in this case too, the DNN model only approximates the desired test distribution, as best as it can, and show some bias toward the head classes as shown in the Figure~\ref{fig:learned_priors}. 
We denote the distribution modeled by the inference time model as $P^{\overline{m}}(y|x)$ and corresponding training time adjusted distribution is denoted by $P^{{m}}(y|x)$ as before. 
These two distributions follow the relation shown in the Eq.~\ref{eq:ideal_adj2}. 
We now derive the optimal adjustment for correcting $P^{\overline{m}}(y|x)$. 
\vspace{-0.7em}
\subsubsection*{\textbf{Theorem 2.}}
 Let $P^m(y|x)$ represent the logit adjusted model and $P^{\overline{m}}(y|x)$ represent the model where logit adjustment is removed during inference, then the optimal adjustment is given by,
\begin{align}
P^a(y|x) &= P^{\overline{m}}(y|x)\frac{P^t(y)}{P^{\overline{m}}(y)} \label{eq:pa2}
\end{align}
where, 
\begin{align}
P^m(y|x) &= P^{\overline{m}}(y|x) \frac{P(y)}{P^{t}(y)} \frac{P^{t}(x)}{P(x)} \\
P^{\overline{m}}(y) &= \int P^{\overline{m}}(y|x) P^t(x) dx \label{eq:pm_}
\end{align}
\\\textbf{Proof.} 
We first derive the optimal adjustment $P^a(y|x)$. From Eq.~\ref{eq:pm} we have,
\vspace{-2em}
\begin{align}
P^m(y) &= \int P^m(y|x) P(x) dx \\
    &= \int P^{\overline{m}}(y|x) \frac{P(y)}{P^{t}(y)} \frac{P^{t}(x)}{P(x)} P(x) dx \\
    &= \frac{P(y)}{P^{t}(y)} \int P^{\overline{m}}(y|x) P^{t}(x) dx \\
    &= \frac{P(y)}{P^{t}(y)} P^{\overline{m}}(y) \label{eq:pm2pmbar}
\end{align}
Substituting the definitions of $P^m(y|x)$ and $P^m(y)$ in Eq.~\ref{eq:pa}, we have,
\begin{align}
P^a(y|x) &= P^{\overline{m}}(y|x) \frac{P(y)}{P^{t}(y)} \frac{P^{t}(x)}{P(x)} \frac{P^t(y)}{\bigg(\frac{P(y)}{P^{t}(y)} P^{\overline{m}}(y)\bigg)}\frac{P(x)}{P^t(x)} \\
    &= P^{\overline{m}}(y|x) \frac{P^{t}(y)}{P^{\overline{m}}(y)} 
\end{align}
Thus the marginal class probabilities on the test distribution are given by, 
\begin{align}
P^a(y) &= \int P^a(y|x) P^t(x) dx\\
    &= \frac{P^t(y)}{P^{\overline{m}}(y)} \int P^{\overline{m}}(y|x) P^t(x) dx\\
    &= P^t(y)
\end{align}

\subsection{Estimating the effective model prior}
So far we have shown that it is possible to adjust the probabilities of the plain cross-entropy trained model as well as training time logit adjusted model to mitigate the model bias. 
Theorem 1 and 2 show the corresponding optimal adjustment such that effective prior of the adjusted model on the test data is $P^t(y)$. 
However, the adjustment depends on $P^m(y)$ and $P^{\overline{m}}(y)$ and we need to estimate these terms accurately. 
We estimate $P^m(y)$ by numerically approximating the integral over the training dataset as, 
\begin{align}
P^m(y) &\approx \frac{1}{\sum_k n_k} \sum_{x \in P(x)} P^m(y|x)\label{eq:pm2}
\end{align}

Similarly, we estimate $P^{\overline{m}}(y)$ using samples from $P^t(x)$ as shown below. 
\begin{align}
P^{\overline{m}}(y) &\approx \frac{1}{\sum_k n_k} \sum_{x \in P^t(x)} P^{\overline{m}}(y|x)\label{eq:pm_2}
\end{align}

We use hold-out validation dataset to represent $P^t(x)$. 
One may note that, in general a large amount of data is needed to have an accurate estimates of Eq.~\ref{eq:pm2} \&~\ref{eq:pm_2}. 
Although, the train dataset size is large enough, validation datasets are limited in size and hence the estimates of $P^{\overline{m}}(y)$ are relatively inaccurate. 
However, we can leverage the training dataset and improve the estimate of $P^{\overline{m}}(y)$. 
In particular, combining Eq.~\ref{eq:pm2pmbar} \&~\ref{eq:pm2} we have, 
\begin{align}
P^{\overline{m}}(y) &\approx \frac{1}{\sum_k n_k} \sum_{x \in P(x)} P^m(y|x) \frac{P^{t}(y)}{P(y)} \label{eq:pmbar2pm}
\end{align}
Thus $P^{\overline{m}}(y)$ can be estimated more reliably using training data itself. 
In practice, we take both estimates of $P^{\overline{m}}(y)$ and average the result which improves the performance further as shown in the ablation experiments. 
Further, similar to~\cite{menon2020long} we tune a scalar hyper-parameter $\alpha$ and use the scaled estimates. 
As the terms $P^{{m}}(y)$ and $P^{\overline{m}}(y)$ represent the effective priors of the trained model and it is used in the adjustment of the respective posterior distributions $P^{{m}}(y|x)$ and $P^{\overline{m}}(y|x)$, we term the proposed post-hoc adjustment as Prior to Posterior: P2P.

\begin{table}[]
\centering
\addtolength{\tabcolsep}{-0.4em}
\begin{tabular}{@{}l|ccc|c|ccc@{}}
\toprule
             & \multicolumn{3}{c|}{Forward}      & \multicolumn{1}{c}{Uniform} & \multicolumn{3}{|c}{Backward}     \\ \midrule
Imbalance ratio     & 50   & 10   & 5    & 1       & 5    & 10   & 50   \\ \midrule
De-Confound~\cite{tang2020long}         & 64.1 & 60.1 & 57.8 &  52.0    &  45.8 & 43.4 & 38.4 \\
Bal-Soft~\cite{balms}    & 62.5 & 58.8 & 57.0  & 52.3 &  46.5 & 44.1 &  39.7 \\
PC-Softmax~\cite{hong2021disentangling}          & 66.6  & 60.6 & 58.1 &  52.8 &  49.3 & 48.8 & 49.0  \\
LADE~\cite{hong2021disentangling}                & 67.4  & 61.3 & 58.6 &  53.0 &  49.8 & 49.2  & 50.0 \\ \midrule
\textbf{Our FT+P2P }          &  \textbf{67.6} &		\textbf{61.5} &	\textbf{58.7}  &	\textbf{54.67} &	\textbf{51.0} &	\textbf{50.5}  &	\textbf{51.1} \\ \bottomrule
\end{tabular}
\caption{Top 1 Accuracy on test time shifted ImageNet-LT dataset. }
\label{tab:shifted_imnet}
\vspace{-1.65em}
\end{table}

\section{Experiments}
\subsection{Experimental setup}

\textbf{Datasets:} We evaluate our proposed approach on long-tailed recognition datasets with different imbalance factors. 
Imbalance factor for a dataset is defined as a ratio between the number of instances in most frequent and least frequent classes. 
CIFAR100-LT and CIFAR10-LT is derived from the original CIFAR100~\cite{krizhevsky2009learning} and CIFAR10~\cite{krizhevsky2009learning} datasets by down-sampling the training instances.
ImageNet-LT dataset is obtained from the larger ImageNet dataset~\cite{liu2019large}. 
It consists of 1000 distinct classes with training instances varying from $5$ to $1280$ for all classes. 
The iNaturalist18~\cite{van2018inaturalist} is a large-scale imbalanced dataset which highly resembles real world scenario. 
We use the standard splits for training and validation. 
As is customary, we keep a balanced validation/test set for evaluation unless otherwise noted. 
\\
\textbf{Baselines:} We compare our method with several approaches proposed in the literature. 
We include methods which focus on minimizing distribution misalignment between train and test datasets such as~\cite{balms,cao2019learning,hong2021disentangling,wang2023margin,zhang2021distribution}, approaches focusing on loss modifications to address imbalanced data training such as~\cite{cui2019class,kang2019decoupling,samuel2021distributional,alshammari2022long}. 
Additionally, we also include some of the recently proposed methods~\cite{cui2022reslt,peifeng2023feature,nam2023decoupled,chen2023area,zhou2023class,ma2023curvature}. 

\noindent
\textbf{Implementation Details:} For CIFAR100-LT and CIFAR10-LT we use ResNet32 as a backbone as introduced in~\cite{cao2019learning}. For ImageNet-LT and iNaturalist18 experiments we use ResNext50 (x50) and ResNet50 (r50) backbone architectures, respectively. 
We train the backbone architecture for $20K$ iterations for both CIFAR datasets with batch size of $64$. 
For ImageNet-LT and iNaturalist18 datasets, we train the backbone for $200$ epochs with batch size of $64$ following~\cite{alshammari2022long}. 
We follow the two stage decoupled training approach. 
In Stage 1, for all cases, we follow common training protocol and use softmax cross entropy loss with instance balanced sampling. 
During Stage 2, we use logit adjusted balanced softmax loss~\cite{balms}. 
We provide further implementation details below. 
\\
\textbf{Classifier Retraining (CL):}
In this case, we initialize the feature backbone with weights from stage 1 and randomly initialize the classifier layer. 
We freeze the backbone model and train only the classifier layer. 
\\
\textbf{Feature Tuning (FT):} 
In these experiments, we use the same settings as used for classifier retraining except that we train the whole model end to end. 
\\
\textbf{Post-hoc adjustment with learned prior (P2P):}
For post-hoc logit adjustment, there is no training involved. 
We estimate the prior using Eq.~\ref{eq:pm2} \&~\ref{eq:pm_2} for respective cases using outputs of the trained model. 

For P2P adjustments, we tune $\alpha$ using holdout validation dataset and report results with tuned $\alpha$, unless otherwise noted. 
We show the results with P2P adjustment for 3 cases, Stage 1 model trained with plain cross entropy loss (CE) and adjusted as per Eq.~\ref{eq:pa} and two of the Stage 2 models trained with logit adjusted loss (classifier training (CL) and whole model training (FT) cases) adjusted as per Eq.~\ref{eq:pa2}. 
We follow rest of the experimental settings as in~\cite{balms}. 

\section{Results}
\subsection{Results on a toy dataset}
We first present the results on a small toy dataset. 
For this, we consider a binary classification problem with an Isotropic Gaussian distribution. 
We use $10000$ samples with imbalance of $100$ for training. 
We repeat the experiment $100$ times and report the average results. 
We train single layer linear model with naive cross-entropy (CE) loss, class frequency based logit adjustment loss and proposed P2P with a learned prior on the CE trained model. 
We show classification boundaries for different approaches in Figure~\ref{fig:toy}. 
We also show the theoretically optimal Bayes classifier in the figures for reference. 
One may note that the classifier boundaries for P2P, which uses learned prior to re-adjust the boundary is very close to the optimal Bayes classifier. 

\subsection{Result on large scale datasets}
We now compare the effectiveness of proposed approach in comparison to other methods from the literature on CIFAR100/10-LT datasets in Table~\ref{tab:cifar}. 
We show results for imbalance factors of $200$, $100$ and $10$. 
We apply post-hoc adjustment on baseline Stage 1 model trained with naive cross entropy loss using proposed P2P approach and class frequency based approach for comparison. 

From the table, we note that proposed post-hoc adjustment is more effective and outperforms class frequency based adjustment for all datasets. 
One may see from the table that proposed post-hoc adjustment on naive cross entropy loss achieves superior performance and outperforms most of the approaches in the literature. 
The performance is further improved for Stage 2 methods with P2P adjustment and best performance is achieved when whole model is trained with logit adjusted loss and combined with post-hoc method (results shown in the last row of the table). 
\begin{figure}[]
  \begin{minipage}[t]{.5\linewidth}
    \includegraphics[width=\linewidth, trim={0cm, 0.2cm, 0cm, 0cm}, clip]{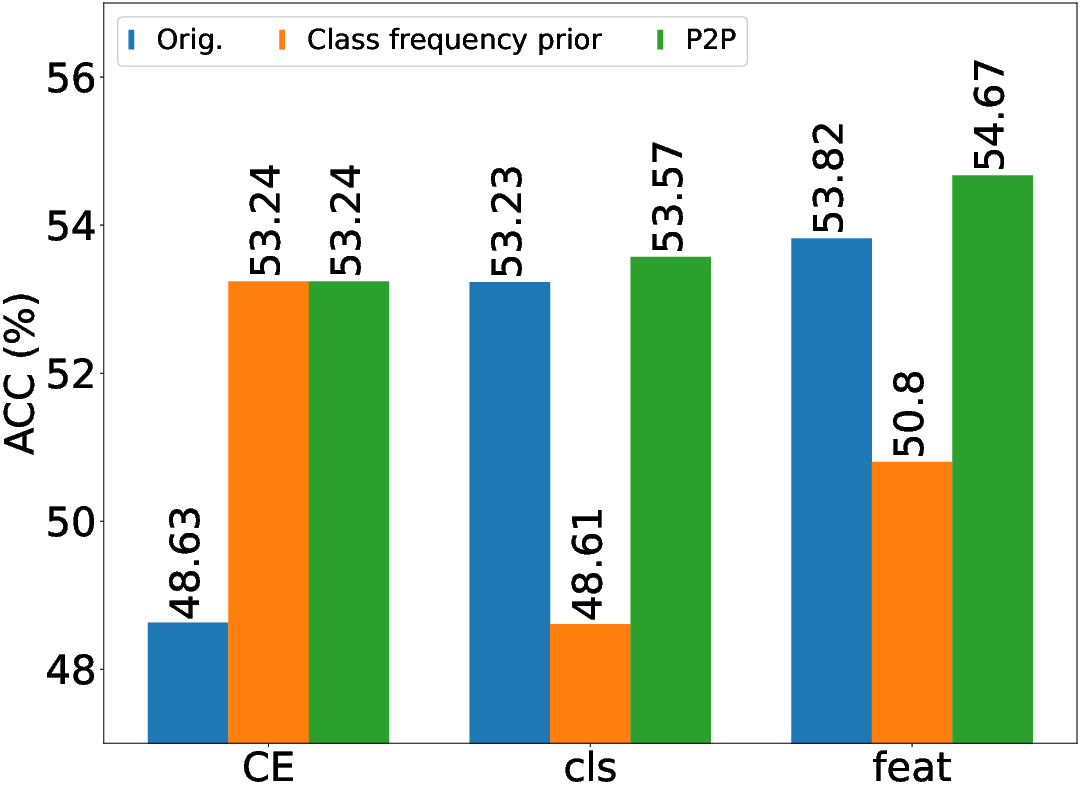}
        \subcaption{\label{fig:priors-approach-imnet}}%
  \end{minipage}\hfill
  \begin{minipage}[t]{.5\linewidth}
    \includegraphics[width=\linewidth, trim={0, 0.5cm, 2cm, 2cm}, clip]{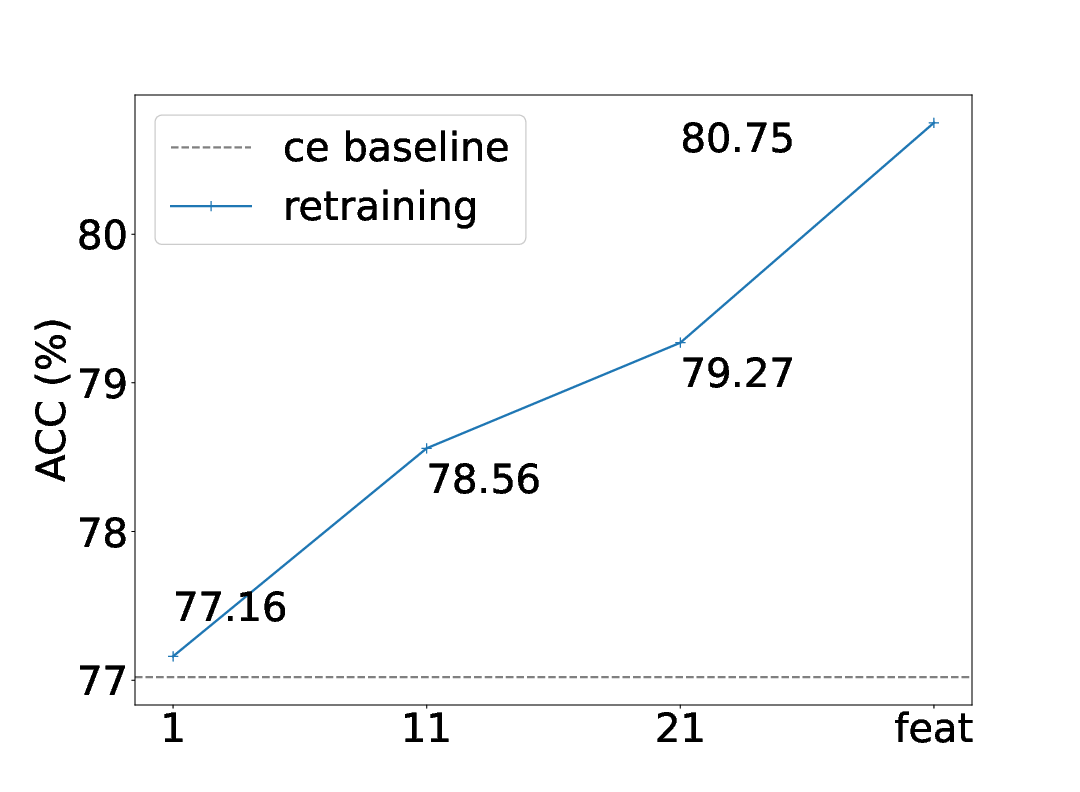}
    \subcaption{\label{fig:ft-ablation}}%
  \end{minipage}%
  \caption{The performance on Imagenet-LT (a) for Stage 1 (CE) and Stage 2 (CL and FT) are shown. The effect of post-hoc using class frequency and P2P can be seen. (b): Accuracy variations with respect to the model depth chosen for fine tuning on CIFAR10-LT dataset with imbalance 200.}
  \vspace{-1.0em}
\end{figure}

We note that using post-hoc adjustment with class frequency bias for Stage 2 models is sub-optimal as most of the data imbalance is removed due to logit adjusted loss. 
However, using proposed approach the residual bias can be calculated and removed. 
We show the results on ImageNet-LT in Figure~\ref{fig:priors-approach-imnet}. 
From the figure one may note that the model performance degrades when using class frequencies for post-hoc correction where as it consistently improves for all methods when using the proposed P2P approach. 
We compare the performance with other methods from the literature in Table~\ref{tab:img-inat-results}. 
Our P2P post-hoc with CE loss achieves $53.24\%$ and $71.15\%$ on ImageNet-LT and iNaturalist18, respectively, already outperforming many SOTA methods. 
Further tuning the classifier and feature backbone and combining it with P2P improves the performance to $53.57\%$ and $54.67\%$ for ImageNet-LT and $71.43\%$ and $71.78\%$ for iNaturalist18, respectively. 
In all the experiments, we notice that the proposed post-hoc approach gives additional performance boost to the model. 

\subsection{P2P on pre-trained models}
One can easily calculate the learned \textit{a posteriori} distribution for many existing pre-trained model and remove the residual bias using P2P. 
Although we explicitly proved that proposed P2P adjustment is optimal for plain CE loss and logit adjusted loss, in principle, this approach can be applied to any method in general. 
In Table~\ref{tab:img_inat-sota} we test this idea on several SOTA methods, including contrastive learning~\cite{zhu2022balanced}, label-noise ~\cite{zhong2021improving}, model ensembles~\cite{wang2020long} and few other methods.  
From the table we see that proposed learned prior consistently improves the accuracy. 
The overall improvement can be as high as $1.48$ and $1.67$ for ImageNet-LT and iNaturalist18 datasets, respectively. These results show that the proposed P2P can be easily integrated with any existing approach to boost the performance. 

\subsection{Evaluation on different imbalances}
In practice, the test dataset can also be imbalanced and moreover, the imbalance of the test dataset can be opposite to that of the train dataset. 
Thus we validate the effectiveness of P2P by simulating different imbalances on the test dataset on ImageNet-LT dataset. 
As shown in Table~\ref{tab:shifted_imnet}, proposed approach consistently performs better in comparison to other methods showing the effectiveness of P2P for different test time distributions. 

\begin{table}[]
\addtolength{\tabcolsep}{-0.5em}
\renewcommand{\arraystretch}{1.2}
\begin{tabular}{@{}c|cc|cc@{}}
\toprule
Dataset      & \multicolumn{2}{c}{ImageNet-LT} & \multicolumn{2}{c}{iNat18} \\ \midrule
Method       & CL            & FT           & CL              & FT              \\ \midrule
Original Acc.   &       53.23        &     53.82         &     70.81            &  71.12               \\\midrule
\multicolumn{5}{c}{\cellcolor[gray]{.9} P2P Adjustment with different $P^{\overline{m}}(y)$} \\ \midrule 
\scriptsize $P_1^{\overline{m}}(y) = \sum_{x \in P(x)} P^m(y|x) P^t(y)/P(y)$ &        53.41       &      54.50        &       71.12          &        71.33         \\
\scriptsize $P_2^{\overline{m}}(y) = \sum_{x \in P^t(x)} P^{\overline{m}}(y|x)$    &  53.35             &      54.38        & 71.11                &  71.37 \\
\scriptsize $P_1^{\overline{m}}(y) + P_2^{\overline{m}}(y)$ &   \textbf{53.57}            &       \textbf{54.67}       &  \textbf{71.43}               &  \textbf{71.78}               \\ \bottomrule
\end{tabular}
\caption{Experiments using different estimates of $P^{\overline{m}}(y)$ using train and validation data samples for post-hoc P2P adjustments. 
}
\label{tab:pmbar-pm}
\vspace{-1.0em}
\end{table}

\subsection{Ablation study}
\textbf{Estimating $P^{\overline{m}}(y)$}
\\
We validate the reliability of estimates $P^{\overline{m}}(y)$ using Eq.~\ref{eq:pm_2} \&~\ref{eq:pmbar2pm} for Stage 2 experiments (CL and FT) on ImageNet-LT and iNaturalist18 datasets. 
As shown in Table.~\ref{tab:pmbar-pm}, the performance improves when P2P is used with either of the estimates for $P^{\overline{m}}(y)$ and the overall best is achieved when both estimates are combined which shows that using more samples improves the accuracy of the estimates. 
\\
\textbf{Is model bias deeply rooted?} 
\\
It is generally argued that the features learned in Stage 1 are sufficient and only classifier boundary adjustment is needed in Stage 2 to remove the bias. 
We check this hypothesis on CIFAR10-LT dataset with imbalance factor of 200. 
In Stage 2, we train different layers in the model with logit adjusted loss. 
We train the ResNet-32 model with 1, 11, 21 and 33 (all) layers, respectively. 
Figure \ref{fig:ft-ablation} show that the performance steadily improves when more layers are trained. 
In our opinion, although Stage 1 with instance balanced sampling generates good quality features, they are tightly clustered for tail classes which might affect their quality~\cite{zhu2022balanced,liu2023inducing}. 
In Stage 2, when classifier boundaries are readjusted, the model can improve upon previous features. 
As a result, feature tuning with post-hoc correction achieves the best performance in all our experiments. 
\section{Conclusions}
We proposed a simple yet effective approach to accurately represent the effective prior of a trained model and use it to mitigate the model bias for long-tailed recognition. 
We present theoretical analysis and prove that proposed approach is optimal for plain cross-entropy loss training and logit adjusted training. 
We show that proposed post-hoc adjustment achieves improved performance and can rival many SOTA methods when combined with classifier or feature retraining. 
Our results show that the proposed approach can be used to mitigate residual bias from existing methods and boost the performance without any need for retraining. 

{\small
\bibliographystyle{ieee_fullname}
\bibliography{sample-base}
}

\clearpage
\setcounter{page}{1}
\maketitlesupplementary

\section{Additional training details}
We train Stage 1 models using cross-entropy loss for CIFAR10-LT and CIFAR100-LT datasets for 20,000 iterations.
In Stage 2, for both classifier and feature tuning cases, we train the models for 4000 iterations. 
For ImageNet-LT dataset, we train the Stage 1 and both the Stage 2 methods for 200 and 20 epochs, respectively. 
For iNaturalist18 dataset, we train the Stage 1 and both Stage 2 models for 200 and 10 epochs, respectively. 
For all models the batch size of 64 is used. 
Rest of the experimental settings are borrowed from~\cite{balms}. 

\section{Dissimilarity of data distributions: $P(X) \neq P^{t}(X)$}
In the paper we clearly maintain the distributions $P(X)$ and $P^{t}(X)$ as distinct in nature. For the sake of completeness we provide a mathematical justification using the approach of moment matching. 
In particular, we show that the first moment of train and test data distribution is not the same. 

\begin{align}
\mu_{x} &= \int_x \int_y x P(x,y) dx dy\\
&= \int_y \left( \int_x x P(x|y) dx \right) P(y) dy\\
&= \int_y \mu_{x|y} P(y) dy\\
\end{align}

where, $\mu_{x|y}$ represents the mean of conditional distribution $P(x|y)$ and $\mu_{x}$ represents the mean of data distribution $P(x)$. 

Similarly we can show that for the test time distributions,
\begin{align}
\mu^t_{x} &= \int_y \mu^t_{x|y} P^t(y) dy\\
\end{align}

Given that the class conditional data distributions $P(x|y)$ and $P^t(x|y)$ are the same, we have,

\begin{align}
\mu_{x|y} = \mu^t_{x|y}
\end{align}
However, since the class prior distributions are not the same by definition, i.e. $P(y) \neq P^t(y)$, we have 

\begin{align}
\mu_{x} \neq \mu^t_{x}
\end{align}
And hence, 
\begin{align}
P(X) \neq P^{t}(X)
\end{align}

\begin{figure}
\centering
\begin{subfigure}{0.45\linewidth}
    \includegraphics[trim={0cm 0cm 0cm 0cm},clip,width=\textwidth]{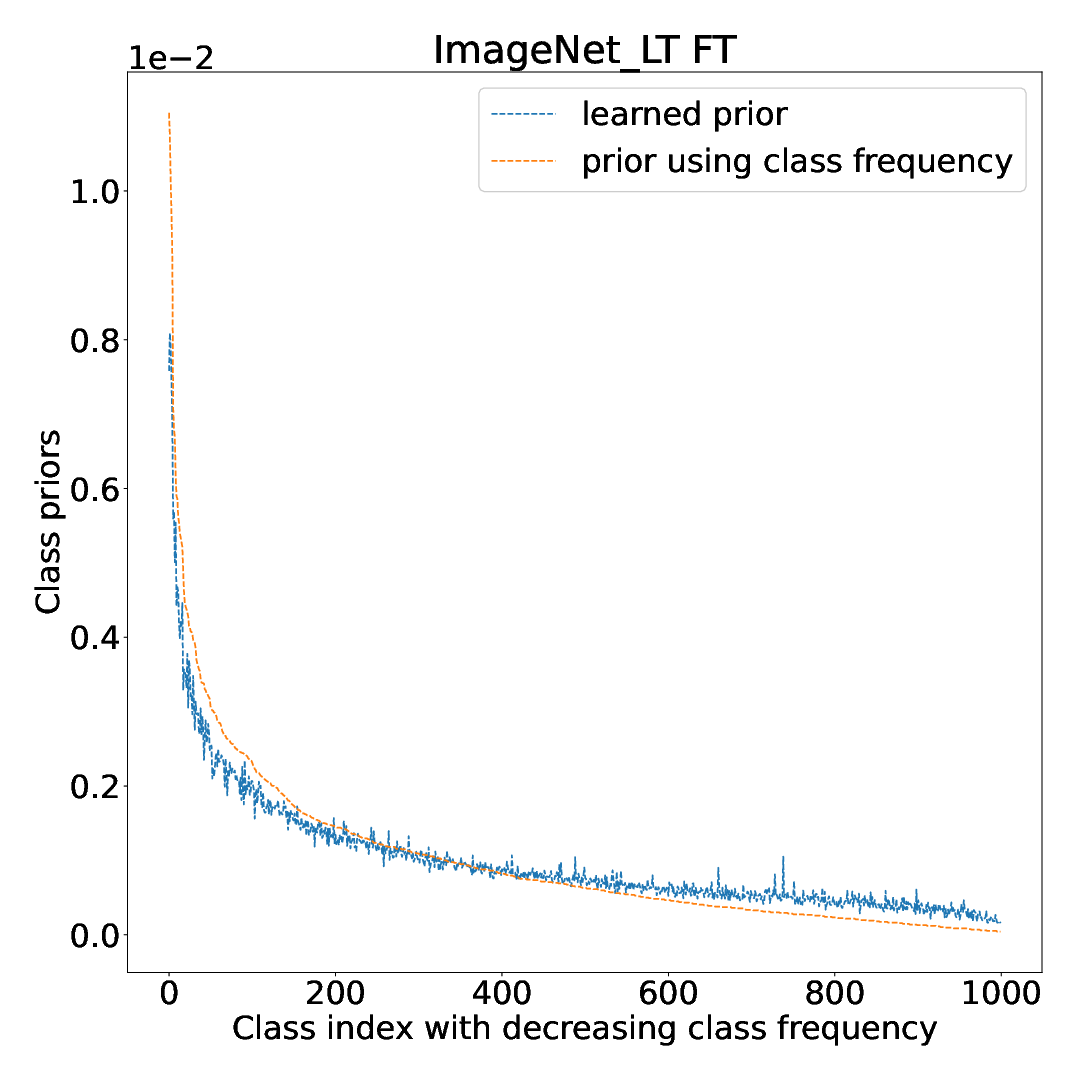}
    \caption{Feature tuning in Stage 2 on ImageNet-LT}
    \label{fig:dp}
\end{subfigure}   
\hfill
\begin{subfigure}{0.45\linewidth}
    \includegraphics[trim={0cm 0cm 0cm 0cm},clip,width=\textwidth]{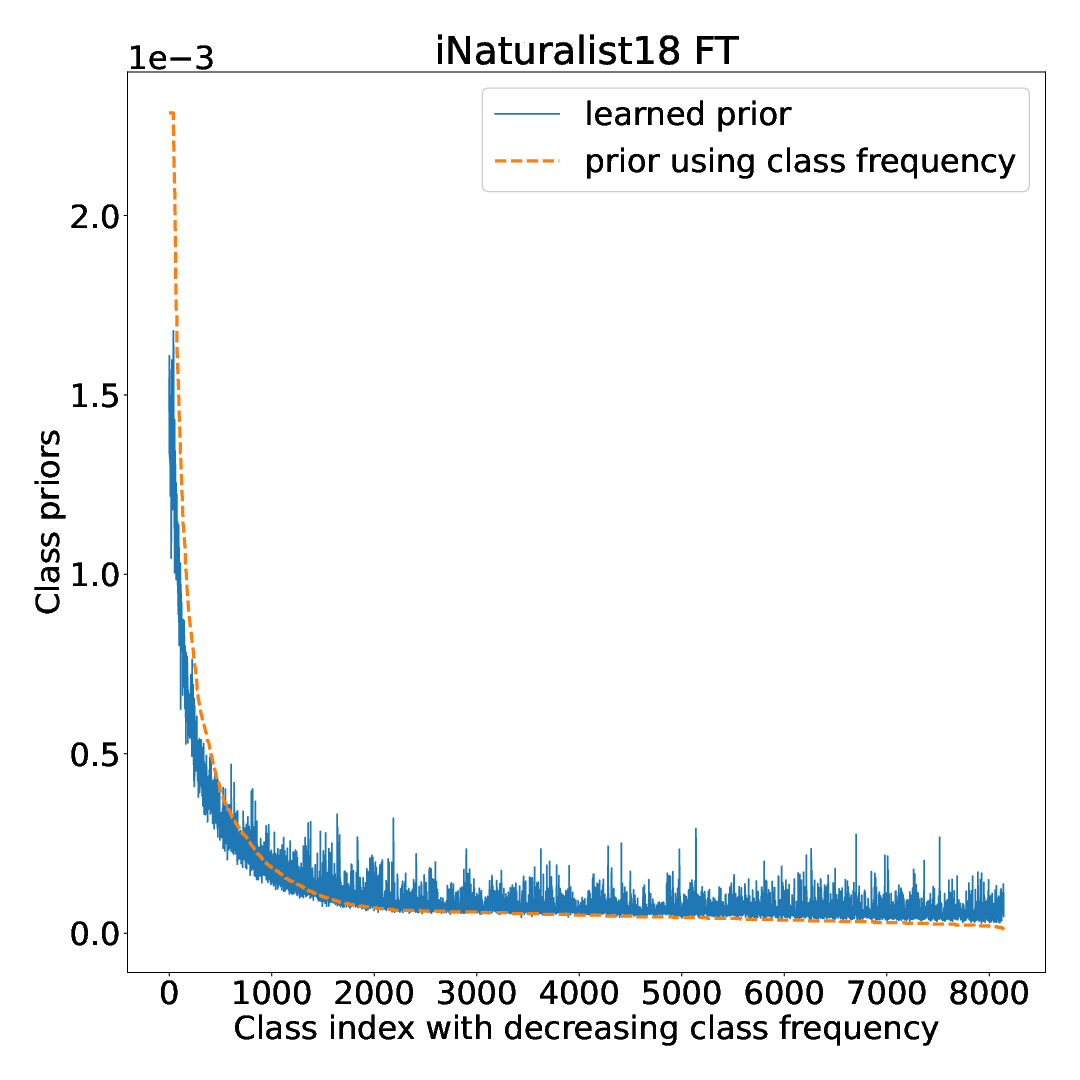}
    \caption{Feature tuning in Stage 2 on iNaturalist18}
    \label{fig:dp}
\end{subfigure}   
\caption{We show model biases for different cases. Bias estimated using class frequencies and proposed method are shown for ImageNet-LT and iNaturalist18 datasets for logit-adjusted feature tuning in Stage 2.}
\label{fig:biases}
\end{figure}

\section{The effective prior for ImageNet-LT and iNaturalist18 datasets}
In Figure~\ref{fig:biases} we show the model bias estimated using class frequencies and the effective prior calculated using proposed approach on ImageNet-LT and iNaturalist18 datasets. 
From the figure it is clearly observed that model bias is quite different from empirical bias estimated using class frequencies. 
In particular, the effective prior is higher for low frequency classes than the class frequency based prior for both datasets.

\begin{table}[h]
\centering
\begin{tabular}{@{}l|cccc@{}}
\toprule
\multicolumn{5}{c}{ImageNet-LT} \\
\hline
Method                  &Many   &Medium &Few    &All \\
\hline

Bal-Soft~\cite{balms}          &64.10  &48.20  &33.40  &52.30\\
RSG~\cite{wang2021rsg}              &63.20  &48.20  &32.20  &51.80\\
LADE~\cite{hong2021disentangling}              &65.10  &48.90  &33.40  &53.00\\
DisAlign~\cite{zhang2021distribution}         &62.70  &52.10  &31.40  &53.40\\
ResLT~\cite{cui2022reslt}             &63.00  &\underline{53.30}  &35.50  &52.90\\
WB+MaxNorm~\cite{alshammari2022long}         &62.50  &50.40  &\textbf{41.50}  &53.90\\
MARC~\cite{wang2023margin}              &60.40  &50.30  &36.60  &52.30\\
RBL~\cite{peifeng2023feature}               &64.80  &49.60  &34.20  &53.30\\
CCL~\cite{tiong2023improving}]               &60.70  &52.90  &\underline{39.00}  &54.00\\
NC-DRW-cRT~\cite{liu2023inducing}            &65.60  &51.20  &35.40  &\underline{54.20}\\

\hline

CE                                      &\textbf{68.11}        &42.56  &14.85  &48.63 \\
CE + P2P             &63.36        &49.99  &36.03  &53.24 \\
\hline
CL                                      &63.85   &49.95  & 34.75 &  53.23 \\
CL + P2P             &62.12   &51.18  & 37.79 &  53.57 \\
\hline
FT                                      &\underline{65.83}   &51.32  & 28.22 &  53.82 \\
FT + P2P             &62.44   &\textbf{53.34}  & 36.06 &  \textbf{54.67} \\
\hline
\end{tabular}
\caption{The table show many, medium and few shot accuracies on ImageNet-LT dataset. Best and Second best results are shown in \textbf{bold faces} and \underline{underlined}.  }
\label{tab:img_shot}
\vspace{-1em}
\end{table}

\section{Multishot accuracies}
\begin{figure}
   \includegraphics[trim={0cm 0cm 0cm 0cm},clip,width=\columnwidth]{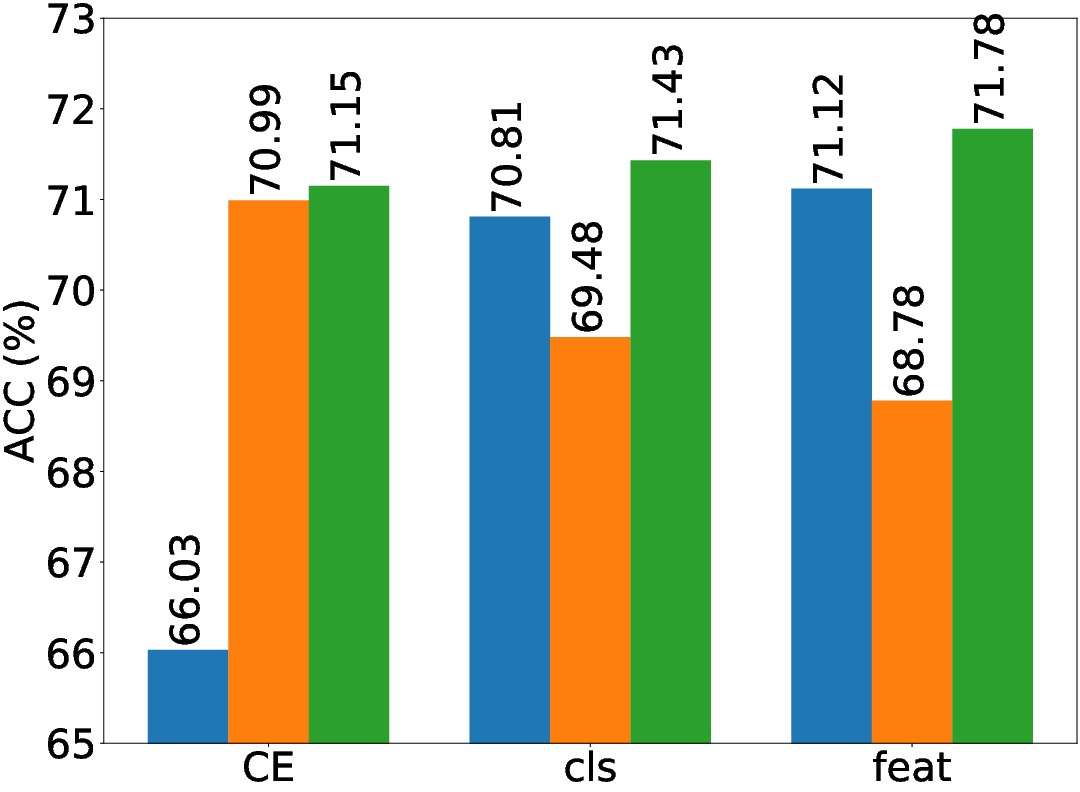}
  \caption{The performance on iNaturalist18 for Stage 1 baseline (CE) and Stage 2 (CL and FT) are shown. The effect of post-hoc using class frequency and proposed approach can be observed.}
  \label{fig:priors-approach-inat}
\end{figure}

In Table~\ref{tab:img_shot} and Table~\ref{tab:inat_shot} we show multi-shot accuracies for ImageNet-LT and iNaturalist18 datasets and compare it with some of the recently published methods. 
We note from the table that proposed approach achieves highest overall accuracy while shot-wise accuracies are not affected much. 
We also show in Figure~\ref{fig:priors-approach-inat} the performance on iNaturalist18 for models trained with plain CE and with logit-adjustment (CL and FT). It can be noted that, P2P outperforms baseline class frequency based adjustment in all the cases.

\begin{table}[t]
\centering
\addtolength{\tabcolsep}{-0.25em}
\begin{tabular}{@{}l|cccc@{}}
\toprule
\multicolumn{5}{c}{iNaturalist18} \\
\hline
Method                 &Many   &Medium &Few    &All \\
\hline

DisAlign~\cite{zhang2021distribution}                   &69.00  &71.10  &70.20  &70.60\\
LDAM+DRW+SAM~\cite{rangwani2022escaping}           &64.10  &70.50  &71.20  &70.10\\
WB+MaxNorm~\cite{alshammari2022long}                 &71.20  &70.40   &69.70   &70.20\\
ResLT~\cite{cui2022reslt}                      &68.50  &69.90  &70.40  &70.20\\
SWA+SRepr~\cite{nam2023decoupled}                  &70.70  &70.83  &70.76  &70.79\\
CC-SAM~\cite{zhou2023class}                     &65.40  &70.90  &72.20  &70.9\\
\hline
CE                      &\textbf{76.33}   &68.15   &60.66   &66.03 \\
CE + P2P      &67.02   &71.05   &\textbf{72.36}   &71.15 \\
\hline
CL                      &70.35   &70.98   &71.06   &70.81 \\
CL + P2P      &68.09   &71.15   &72.19   &\underline{71.43} \\
\hline
FT                      &\underline{71.81}   &\underline{71.46}   &70.16   &71.12 \\
FT + P2P      &66.63   &\textbf{71.73}   &\underline{72.32}   &\textbf{71.78} \\
\hline
\end{tabular}
\caption{The table show many, medium and few shot accuracies on iNaturalist18 dataset. Best and Second best results are shown in \textbf{bold faces} and \underline{underlined}.}
\label{tab:inat_shot}
\end{table}

\begin{table*}[]
\centering
\begin{tabular}{@{}l|ccccc|c|ccccc@{}}
\toprule
             & \multicolumn{5}{c|}{Forward}      & \multicolumn{1}{c}{Uniform} & \multicolumn{5}{|c}{Backward}     \\ \midrule
Imbalance ratio     & 50   & 25   & 10   & 5    & 2    & 1       & 2    & 5    & 10   & 25   & 50   \\ \midrule
CE             & 66.3 & 63.9 & 60.4 & 57.1 & 52.3 & 48.63 & 44.2 & 38.9 & 35.0 & 30.5 & 27.9     \\ \midrule
De-Confound~\cite{tang2020long}         & 64.1 & 62.5 & 60.1 & 57.8 & 54.6 & 52.0    & 49.3 & 45.8 & 43.4 & 40.4 & 38.4 \\
Bal-Soft~\cite{balms}    & 62.5 & 60.9 & 58.8 & 57.0 & 54.4    & 52.3 & 49.6 & 46.5 & 44.1 & 41.4 & 39.7 \\
PC Causal Norm~\cite{hong2021disentangling}      & 66.7 & 64.3 & 60.9 & 58.1 & 54.6 & 52.0 & 49.8 & 47.9 & 47.0 & 46.7 & 46.7     \\
PC-Balanced Softmax~\cite{hong2021disentangling} & 65.5 & 63.1 & 59.9  & 57.3 & 54.3 & 52.1 & 50.2 & 48.8 & 48.3 & 48.5 & 49.0    \\
PC-Softmax~\cite{hong2021disentangling}          & 66.6 & 63.9 & 60.6 & 58.1 & 55.0 & 52.8 & 51.0 & 49.3 & 48.8 & 48.5 & 49.0  \\
LADE~\cite{hong2021disentangling}                & 67.4 & 64.8 & 61.3 & 58.6 & 55.2 & 53.0 & 51.2 & 49.8 & 49.2 & 49.3 & 50.0 \\ \midrule
\textbf{Our FT+P2P }          &  \textbf{67.6} &	\textbf{64.9} &	\textbf{61.5} &	\textbf{58.7} &	\textbf{56.4} &	\textbf{54.67} &	\textbf{52.3} &	\textbf{51.0} &	\textbf{50.5} &	\textbf{50.8} &	\textbf{51.1} \\ \bottomrule
\end{tabular}
\caption{Top 1 Accuracy on test time shifted ImageNet-LT dataset. }
\label{tab:shifted_imnet}
\end{table*}

\section{Additional results on test time shifted imbalance}
In Table.~\ref{tab:shifted_imnet} we compare model performance for test-time shifted distributions with additional baselines and a few more distribution shifts. We note the superior performance of proposed algorithm.

\begin{figure*}
\centering
\includegraphics[width=\linewidth,trim={0cm, 0cm, 13cm, 0cm}, clip]{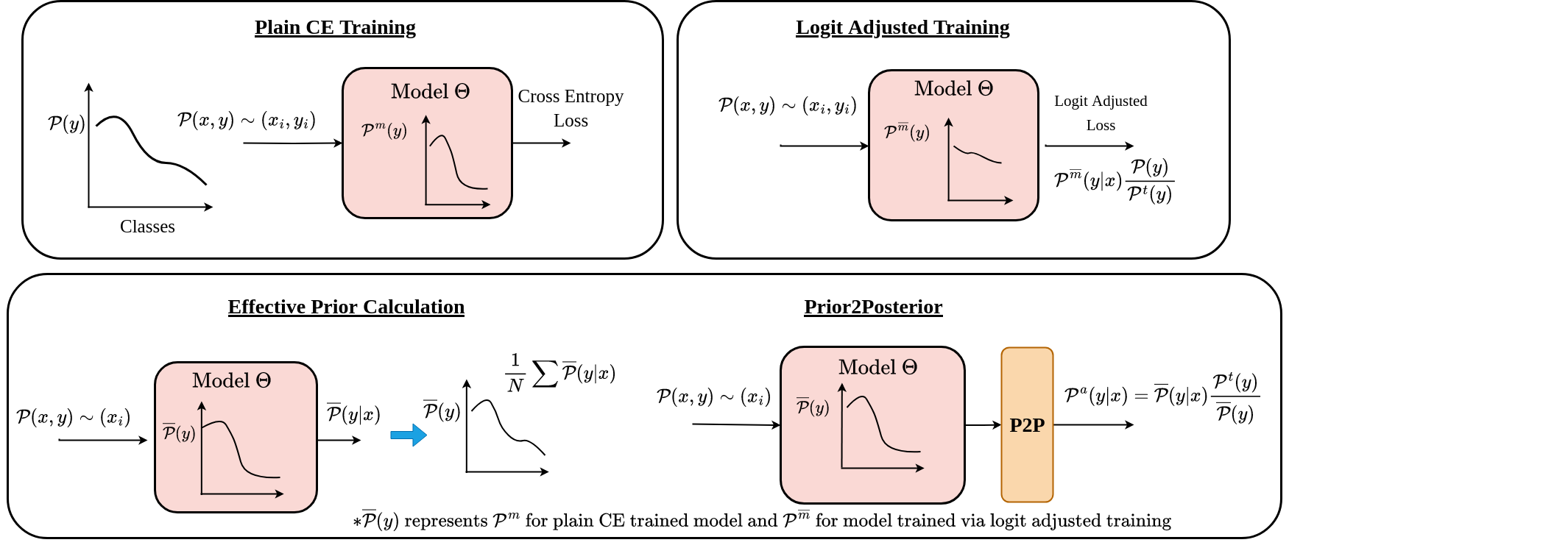}
\caption{ 
Proposed approach is summarized in the figure. 
The top row illustrates plain CE loss and logit-adjusted loss showing that model accumulates some bias due to imbalanced training data. 
We show effective prior calculation for trained model in the bottom row. 
Once the prior is calculated, \textit{a posteriori} probabilities can be corrected using proposed approach as shown in the bottom row. 
}
\label{fig:block_diagram}
\end{figure*}

\section{Discussion on Distribution Matching}
Recent works like~\cite{peng2021optimal} have proposed to tackle this distribution misalignment problem from an optimisation perspective, employing the concept of optimal transport. Although the work provides interesting mathematical insights into relation between the distribution alignment problem and optimal transport, the method assumes that the marginal distribution is consistent with a uniform distribution. Unlike this we impose no such constraint in our proposed approach rendering further flexibility and simplicity in its implementation. 

Similarly~\cite{wang2022solar} also propose optimal transport based distribution matching framework for imbalanced partial label learning. They propose to refine the pseudo-labels in order to align with the true class prior by reducing the optimal transport objective function. 

\cite{shi2024relative} present a novel variant of the optimal transport called, Relative Entropic Optimal Transport to learn matching with a specified prior. The manually specified smoothing guidance matrix $\mathcal{Q}$ can be seen as a generic representation for the effective prior.

\section{Flow of the proposed approach}

We summarise the proposed approach in a block diagram as shown in Figure\ref{fig:block_diagram}. 
The block diagram illustrates the different stages involved in the process starting from bias accumulation in traditional training to bias removal using the proposed method. Both Logit adjusted training and Prior2Posterior  is depicted along with an illustration showing the Effective Prior computation. 


\end{document}